\documentclass[11pt]{article}

\usepackage[preprint]{acl}

\usepackage{times}
\usepackage{latexsym}
\usepackage{amsmath}

\usepackage{booktabs}
\usepackage{caption}
\usepackage{siunitx} 
\usepackage[margin=2.5cm]{geometry} 

\usepackage{spverbatim}
\usepackage{listings}
\usepackage{xcolor}

\usepackage{multirow}
\usepackage{makecell}      
\usepackage{array}         

\usepackage{needspace}

\usepackage{float}

\usepackage[T1]{fontenc}

\usepackage[utf8]{inputenc}

\usepackage{microtype}

\usepackage{inconsolata}

\usepackage{graphicx}

\title{Long-term Task-oriented Agent: Proactive Long-term Intent Maintenance in Dynamic Environments}

\author{
  \textbf{Qinglong Shi\textsuperscript{1}$^*$},
  \textbf{Donghai Wang\textsuperscript{2}},
  \textbf{Hantao Zhou\textsuperscript{2}},
  \textbf{Jiguo Li\textsuperscript{2}},
  \textbf{Jun Xu\textsuperscript{2}\textsuperscript{\dag}}, \\
  \textbf{Jiuchong Gao\textsuperscript{2}\textsuperscript{\dag}},
  \textbf{Jinghua Hao\textsuperscript{2}},
  \textbf{Renqing He\textsuperscript{2}} \\
  \textsuperscript{1}School of Management, University of Science and Technology of China, Anhui, China \\
  \textsuperscript{2}Meituan, Beijing, China \\
  ivyinautumn@mail.ustc.edu.cn \\
  \{xujun58, gaojiuchong\}@meituan.com
}

\begin{document}
\maketitle
\begingroup
  \renewcommand{\thefootnote}{} 
  \begin{NoHyper}
    \footnotetext{$^*$Work done during internship at Meituan}
    \footnotetext{\textsuperscript{\dag}Corresponding authors: Jun Xu, Jiuchong Gao}
  \end{NoHyper}
\endgroup
\begin{abstract}
Current large language model agents predominantly operate under a reactive paradigm, responding only to immediate user queries within short-term sessions. This limitation hinders their ability to maintain long-term user's intents and dynamically adapt to evolving external environments. In this paper, we propose a novel interaction paradigm for \textbf{proactive Task-oriented Agents} capable of bridging the gap between relatively static user's needs and a dynamic environment. We formalize proactivity through two key capabilities, (i) Intent-Conditioned Monitoring: The agent autonomously formulates trigger conditions based on dialog history; (ii) Event-Triggered Follow-up: The agent actively engages the user upon detecting useful environmental updates. We introduce a high-quality data synthesis pipeline to construct complex, multi-turn dialog data in a dynamic environment. Furthermore, we attempt to address the lack of evaluation criteria of task-oriented interaction in a dynamic environment by proposing a new benchmark, namely \textbf{ChronosBench}. We evaluated some leading close-source and open-source models at present and revealed their flaws in long-term task-oriented interaction. Furthermore, our fine-tuned model trained using synthetic data for supervised learning achieves a task completion rate of 85.19\% for complex tasks including shifts in user intent, outperforming other models under test. And the result validated the effectiveness of our data-driven strategy.
\end{abstract}

\section{Introduction}

Large Language Models (LLMs) have achieved remarkable success in serving as conversational assistants \citep{AgentPotentialSurvey}, demonstrating exceptional proficiency in instruction following and knowledge retrieval \citep{AgentCognitive, MultimodalInteraction}. However, the prevailing interaction paradigm remains predominantly reactive \citep{AgentExploration} and session-bound. In this standard setting, the agent functions passively, keeping dormant until triggered by a user query. This paradigm is fundamentally misaligned with real-world scenarios where user needs are often temporal and contingent on dynamic external factors. For instance, a user seeking a specific product or a job may not find an immediate match. As a result, proactive agents have become an increasingly important subject of research.

To bridge this gap, there is a growing consensus on the need to transition towards proactive agents managing tasks \citep{ProactiveTaskManagement, ProactiveAgent}. Some researchers pay attention to equipping conversational agents with proactive interaction abilities \citep{ProactiveConversationalAgent}. Additionally, user-centered agents have been emphasized by many people \citep{UserRL, UserBench, ProactiveEval}. However, existing work has not yet covered agents' ability to engage in long-term task-oriented interaction within dynamic environments.

We believe that proactive agents, beyond differing from their passive counterparts in interaction paradigms, must also demonstrate the ability to \textbf{sustain long-term intentions across discrete timelines}. This proactivity can be defined as a dual capability of \textbf{intent-conditioned monitoring} and \textbf{event-triggered engagement}. Specifically, an effective agent must possess the foresight to convert a user's unfulfilled request into a structured monitoring task and the judgment to reengage the user when the external environment updates to a state that satisfies the conditions. This shift requires the model not only continuously maintain task's state updates and keep monitoring environmental states, but modify triggers of reminding user of important information based on the user's evolving preferences during intermittent interactions \citep{ProPerSim}.

In this work, we propose a comprehensive framework to endow LLMs with these capabilities through a data-driven strategy. We introduce a high-quality data synthesis pipeline that leverages an iterative "generate-and-evaluate" mechanism to construct a conversational training set including totally 1,052 cases across 3 scenarios, which can be described as training set of Table \ref{tab:ChronosBench Dataset Distribution}. We fine-tune base models including Qwen3-8B, Qwen3-32B and Llama-3.1-8B-Instruct to master this interaction paradigm, focusing on enhancing the timing of trigger setting and the context-awareness \citep{Contextagent} of proactive notifications.

To evaluate these capabilities, we construct \textbf{ChronosBench}, a novel benchmark designed to simulate iterative, cross-time interactions with a constantly updating external information stream. Distinct from static benchmarks, our evaluation environment incorporates a simulated timeline where external states evolve constantly. Besides, we categorize test samples into simple and complex tiers. Simple scenes test basic retrieval and reminder setting capabilities while complex scenes introduce intention shift, requiring the agent to detect changes in user needs, update existing triggers, and execute multi-stage proactive follow-ups(or keep silent when need). The overall distribution of the test set can be described as test set of Table \ref{tab:ChronosBench Dataset Distribution}. The evaluation result on ChronosBench strongly demonstrates the effectiveness of our data-driven strategy. For instance, Claude-sonnet-4 achieves a task completion rate of 72.22\% on complex scenes. In contrast, the evaluation result of our fine-tuned Qwen3-32B is 85.19\% on complex scenes, outperforming all close-source and open-source models under test. This lays a solid foundation for evaluating agents' performances in more complex and realistic interaction under relevant scenarios in the future.

\begin{table}[ht]
\centering
\small
\begin{tabular}{>{\raggedright\arraybackslash}m{2.6cm} m{0.45cm} m{0.45cm} m{0.45cm} m{0.45cm} m{0.65cm}}
\toprule
\multirow{2}{*}{\textbf{Scenario}} & 
\multicolumn{2}{c}{\textbf{Simple}} & 
\multicolumn{2}{c}{\textbf{Complex}} & 
\multirow{2}{*}{\textbf{Count}} \\
\cmidrule(lr){2-3} \cmidrule(lr){4-5}
& {Pos.} & {Neg.} & {Pos.} & {Neg.} \\
\midrule
\multicolumn{6}{c}{\textit{\textbf{Training Set}}} \\
\midrule
\textit{Product Recommend}    & 131 & 123 & 52 & 32 & 338 \\
\textit{Job Search}   & 143 & 143 & 46 & 51 & 383 \\
\textit{Flight Booking}  & 127 & 124 & 37 & 43 & 331 \\
\textbf{Total}   & \textbf{401} & \textbf{390} & \textbf{135} & \textbf{126} & \textbf{1,052} \\
\midrule
\multicolumn{6}{c}{\textit{\textbf{Test Set}}} \\
\midrule
\textit{Car Purchase}    & 27 & 27 & 9 & 9 & 72 \\
\textit{House Hunting}   & 27 & 27 & 9 & 9 & 72 \\
\textit{Ticket Booking}  & 27 & 27 & 9 & 9 & 72 \\
\textbf{Total}   & \textbf{81} & \textbf{81} & \textbf{27} & \textbf{27} & \textbf{216} \\
\bottomrule
\end{tabular}
\caption{Training set and Test set of the ChronosBench. The reason why the proportion of the positive to the negative is not exactly 1:1 is that some unqualified samples were filtered out during post-processing. Each sample will branch into positive and negative pathways.}
\label{tab:ChronosBench Dataset Distribution}
\end{table}

\section{Related Work}

\subsection{Proactive Agent Framework}

Recent research has increasingly sought to transition LLM-based agents from passive responders to proactive assistants capable of foresight. The innovation of CollabLLM \citep{CollabLLM} is a collaborative simulation that estimates the long-term contribution of responses using Multiturn-aware Rewards. ContextAgent \citep{Contextagent} is the first context-aware proactive agent that incorporates extensive sensory contexts surrounding humans to enhance the proactivity of LLM agents. With a large-scale dataset with detailed annotations, SalesBot \citep{SalesBot} focuses on investigating the conversations starting from open-domain social chatting and then gradually transitioning to task-oriented purposes. A seminal work in this direction is ProactiveAgent \citep{ProactiveAgent}, which fundamentally challenges the reactive paradigm by endowing agents with the ability to anticipate and initiate tasks without explicit human instruction.

\subsection{Evaluation For Proactivity}

About evaluation of agent's conversational ability, NVIDIA proposed RULER \citep{Ruler} aiming at revealing the true performance of long context models when facing complex instructions. Longbench \citep{Longbench} and its improved version Longbench-v2 \citep{Longbench-v2} have jointly driven the evolution of large models from merely processing long texts to effectively understanding and reasoning about long texts. As agent's capabilities expand, evaluating their ability to align with user needs and guide conversations has become critical. UserBench \citep{UserBench} proposes a user-centric benchmark evaluation framework that interacts with agents through multi-round operations driven by user preferences. In parallel, ProactiveEval \citep{ProactiveEval} proposes a unified evaluation framework that decomposes proactive dialog into target planning and dialog guidance. Introducing dual-control mechanism, $\tau^2-Bench$ \citep{tau2Bench} let both agent and user make use of tools to act in a shared, dynamic environment that tests both agent coordination and communication. 

Our research for agent's proactivity lays emphasis on strengthening the ability of \textbf{long-term user's intent maintenance in a dynamic environment} because the external information may not always satisfy user's specific changeable needs. Compared with the above benchmarks operating largely within the bounds of a continuous interaction session, our work differs by introducing the \textbf{dimension of time and environmental dynamics}. We argue the task-oriented agent must monitor external information over extended periods and adapt to intention shifts that occur during the interaction. Such scenario is not fully covered by previous benchmarks.

\section{Methodology}

In this section, we formalize the problem of proactive intent maintenance and present our proposed framework. We first define the hybrid interaction mechanism that decouples environmental monitoring from agent reasoning. Then, we detail our iterative dialog data synthesis pipeline designed to construct high-quality, long-term interaction including the complex type featuring intention shifts.

\subsection{Problem Formulation}

We model the interaction between a user $U$ and an agent $A$ over a continuous timeline $T$ \citep{ConversationInfo}. At any time step $t$, the environment state is denoted as $E_t$ (e.g., product prices, flight status).
Unlike traditional reactive agents that compute a response $R_t$ solely based on the current user query $Q_t$ and dialog history $H_t$, which can be formalized as:

\begin{equation}
R_t = f(H_t, Q_t, E_t) 
\end{equation}
a proactive task-oriented agent must operate in two distinct modes:

\subsubsection{Intent Maintenance (Reactive Phase)}

Generally speaking, users typically follow this progressive process when expressing their intents: (i) Initially, they present vague, rough requirements \citep{UserBench}; (ii) They then add constraints to refine and specify those requirements; (iii) Sometimes there will be intention shifts in more complex scenarios. Upon receiving a initial user inquiry $Q_t$, the agent must initialize the task description $T_t$ and continuously update it with subsequent supplementary constraints $C$ and intention shifts $S$, which can be formalized as:

\begin{equation}
T_t = g(H_t, Q_t, C, S) 
\end{equation}
To demonstrate agent's state-tracking ability, task descriptions are presented in the form of structured state slots, including intention description $T_d$, constraints $T_c$ and task status $T_s$ (i.e., $T_t = \{T_d, T_c, T_s\}$). When dynamic external information does not meet the user's current needs, the agent shall set reminders based on a thorough understanding of the user's intentions and constraints, so as to proactively notify the user when future external information presents options that meet their requirements. The reminder $M_t$ can be described as:

\begin{equation}
M_t = h(H_t, Q_t, C, S, E_t) 
\end{equation}
We describe the reminder structure as $M_t = \{type, value\}$, of which \textit{type} is the trigger type such as time and event, and \textit{value} is the trigger condition description.

\subsubsection{User Wake-up (Follow-up Phase)}

During the dormancy period where no user input exists ($Q_t = \emptyset$), if the environment state shifts from $E_t$ to $E_{t+\Delta}$, the agent must autonomously generate a wake-up signal and proactively initiate a response $R_{t+\Delta}$ to inform the user when latest environment state satisfies user's requirements. On the contrary, the agent should keep silent when latest environment state still fails to meet user's requirements. 

\subsection{Hybrid-Triggered Proactive Framework}

\begin{figure*}[t]
    \centering
    \includegraphics[width=\textwidth]{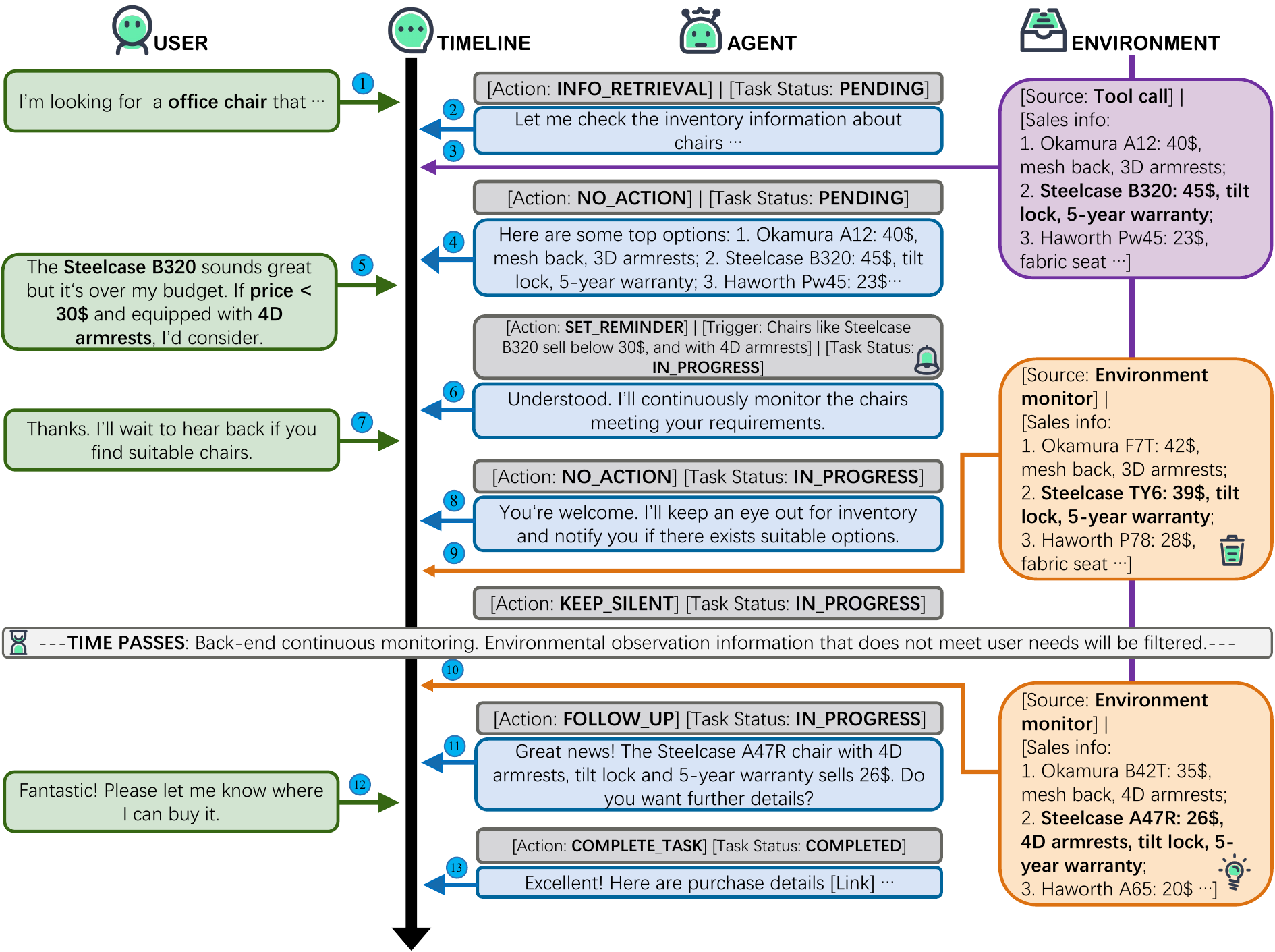}
    \caption{Example of overall task-oriented simple dialog design. According the timeline, the process can generally be divided into three parts: (a). 1st-8th, User active period; (b). 9th-10th, User dormant period; (c). 11th-13th, User wake-up period.}
    \label{fig:Simple Dialog Example}
\end{figure*}

To implement the above formulation efficiently, we design a hybrid information retrieval architecture that unifies synchronous and asynchronous interactions.

The agent serves as a standard assistant, but responds in structured format including a field named \textit{proactive\_action} trained to indicate its current action and a field \textit{response\_text} containing natural language responses. During active dialog sessions, the agent can perform the action \textit{INFO\_RETRIEVAL} to trigger a real-time retrieval of $E_t$ and immediately receive an \textit{observation} message involving retrieval results like ReAct \citep{ReAct}. Strictly speaking, this function-calling-like retrieval method ignores standard process defined by openAI \citep{openai} including predefined tools' JSON Schema registration and previous function calling step \citep{FunctionCalling}, so such observation message is more like externally injected information achieved through engineering methods instead of real results of standard function calling.

We assume a backend monitor that periodically scans the environment state. This module acts as a filter, executing the logic defined by the agent's previous reminders. Crucially, the system constructs an \textit{observation} message to replace the typical user query, and the message contains the updated environment state $E_{t}$ as well as the original task context. This design allows the agent to perceive environmental changes as \textit{internal triggers} without requiring user input.

\subsection{Iterative Data Synthesis Pipeline}

We propose an automated, iterative data synthesis pipeline. To ensure the generated dialogs are logical and not merely distilled from a stronger model, we introduce a Multi-Agent Simulation with Quality Critics. The pipeline consists of three steps:

\textbf{Step 1: Scenario Background Initialization}. We firstly prompt (Refer the prompt in Appendix \ref{sec:appendix Scenario Background Data Generation}) GPT-4.1 to randomly synthesize a diverse set of elements involving user personas, user initial needs, supplemental constraints and multi-stage environmental states on the basis of scenario templates consisting of \textit{product recommendation}, \textit{job search} and \textit{flight booking}. For complex cases, there is an additional element of user's intention shift. Refer the example in Appendix \ref{sec:appendix Scenario Background Data Example} to see the final result.

\textbf{Step 2: Interactive Dialogue Generation}. Based on the existing scenario background, a user simulator is prompted (Refer the prompt in Appendix \ref{sec:appendix User Simulator Prompt}) to engage in an iterative conversation with a task-oriented agent simulator (Refer the prompt in Appendix \ref{sec:appendix Basic Prompt}). All user or agent dialogs are evaluated by a quality controller (Refer two prompts in Appendix \ref{sec:appendix User Dialog Quality Control Prompt} and \ref{sec:appendix Agent Dialog Quality Control Prompt}) played by strong models that can align with human preferences to a considerable degree \citep{MT-Bench}. Only dialogs scoring above a certain threshold are added to the message history list. The user simulator, task-oriented agent simulator and their respective quality controllers are each played by GPT-4.1 with four distinct prompts.

\textbf{Step 3: Double Dialog Branches}. The final conversation is generated sentence by sentence, with the dialog end marker determined by the sample branch type. For positive samples where the updated environmental state satisfies the user's needs, the agent should indicate task completion. For negative samples, the agent must ultimately remain silent. Refer the examples in Appendix \ref{sec:appendix Complex Conversation Data Example}.

\subsection{Task-oriented interactive paradigm}

During the design of a dialog dataset pipeline for fine-tuning base models, we have the following basic assumptions: (i) User expresses their intent progressively, adding constraints to enrich the details of their initial intent; (ii) The initial environmental state information cannot immediately meet the user's initial requirements including limitations; (iii) During user's dormancy, the back-end monitoring framework delivers the latest environmental status information to the agent for processing at a fixed frequency. Based on the above assumptions, we can simply describe the conversation(for simple cases) like the content depicted in the Figure \ref{fig:Simple Dialog Example}. For the complex scenes involving user's intent shift, there are more assumptions: (iv) User will only express their intent shift after the agent first follow-up with information meeting their initial requirements; (v) The firstly updated environment state cannot immediately meet the user's changed requirements so that there is a need for the agent setting a reminder again. Based on the additional assumptions, we can describe the conversation(for complex cases) like the content depicted in the Figure \ref{fig:Complex Dialog Example}.

Besides above assumptions, we construct the task-oriented agent's interactive behavior through structured JSON-formatted responses. In addition to presenting natural language output to users like a typical intelligent assistant, the agent communicates with the system back-end internally through the `proactive\_action` field described in Appendix \ref{sec:appendix Task-oriented Agent's Available Actions}.

It should be pointed out that, setting up a reminder essentially creates a task within a database-driven tracker module, involving data like username, task's conditions, task status, and last interaction time. The specific design approach for the overall task-oriented agent system architecture falls within the realm of engineering implementation methods. This is not directly relevant to the theme explored in this paper, so we will not elaborate further here.

\subsection{Intention Shift}

\begin{figure*}[t]
    \centering
    \includegraphics[width=\textwidth]{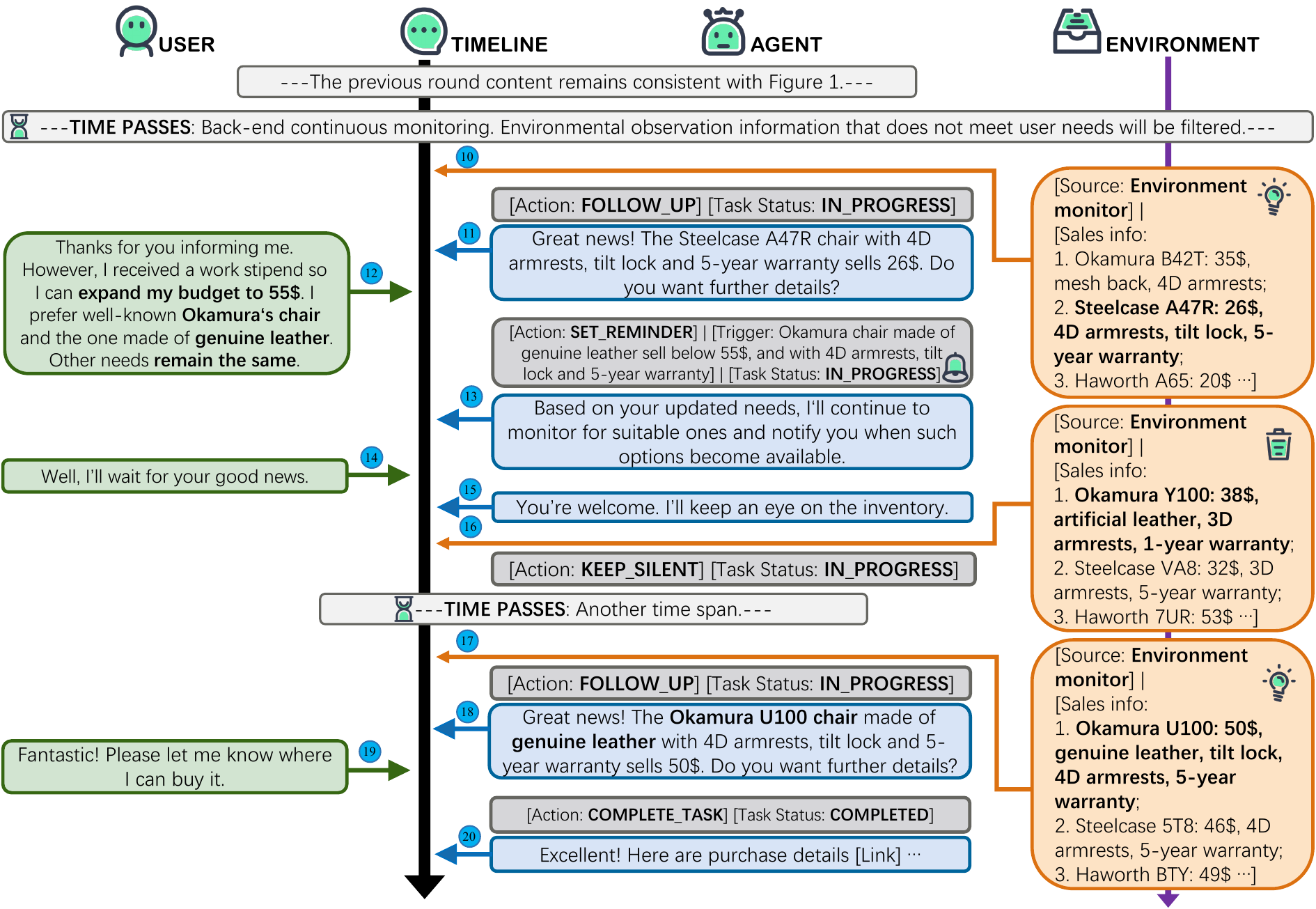}
    \caption{Example of complex dialog involving the intention shift. According the timeline, the process can generally be divided into three parts: (a). 11th-15th, User first wake-up period; (b). 16th-17th, Another user dormant period; (c). 18th-19th, Another user wake-up period.}
    \label{fig:Complex Dialog Example}
\end{figure*}

To simulate as closely as possible the characteristic of human intent changing over time in the real world, we have introduced a mechanism named \textit{Intention Shift}. In the simple interaction example described as Figure \ref{fig:Simple Dialog Example}, though the user's initial intent may have been refined with additional constraints or details in subsequent rounds, these changes still occurred within a relatively short time frame. We assume that users may express that their intentions have changed when the agent first proactively provides information that meets their initial requirements. Generally speaking, the shifted intent does not completely deviate from the original one. Most of the time, the core elements remain unchanged(e.g., still need a chair but add new features), with only additional constraints and supplementary requirements being added. As depicted in Figure \ref{fig:Complex Dialog Example}, we temporarily assume that user will immediately become active and respond with their intent shift after the agent's first follow-up. The assumption means that, during the agent's first follow-up and the user's intent shift, the environmental state remained largely unchanged. Consequently, it's better for the agent to set a new reminder immediately after user describes their intention shift rather than performing yet another redundant information retrieval by `INFO\_RETRIEVAL`. Of course, there's nothing wrong with doing so.

\subsection{Task State Tracking}

The task-oriented agent needs to keep updated with the task process by continuously modifying the `status` field in the `task\_description` field (`task\_description` is stored in JSON format). As for the status of the task, available options are "PENDING", "IN\_PROGRESS", "COMPLETED" and "FAILED". When the agent has not set any reminders in current dialog, the status should be "PENDING". When at least one reminder has been set and the task is not completed, the status should be "IN\_PROGRESS". When user explicitly expresses that their needs have been met (perhaps express gratitude), the status should be "COMPLETED". When user expresses refusal/disappointment or asks for cancellation of the task, the status should be "FAILED".

\section{Experiments}

In order to obtain the Proactive Task-oriented Agent, we use the training set of ChronosBench to train three open-source models: Qwen3-8B, Qwen3-32B and LLaMA-3.1-8B-Instruct. For training Qwen3-32B, we employ LoRA rank of 32, a total batch size of 16, a learning rate of $3e - 5$, and an AdamW Optimizer with a 0.1 warm-up ratio. We train the model for 3 epochs and use 2 A100 GPUs on one node to train for approximately 2 hours. For training Qwen3-8B and LlaMA-3.1-8B-Instruct, we employ LoRA rank of 16, a total batch size of 16, a learning rate of $5e - 5$, and Optimizer remains the same as well as warm-up ratio. We train the two models for 5 epochs and the process takes 1 hour or so respectively with similar computing resources. Additionally, we set the template for formatting training data to `qwen3\_nothink` so that the thinking models like Qwen3 series respond in single structured JSON format. The detailed prompt can be found in Appendix B. The automatic evaluation of these metrics relies on comprehensively designed evaluation scripts. The models under test engage in dialogs with the user simulator on the test set and the temperature is set to 0.2.

\subsection{Metrics}

\begin{table*}[t]
\centering
\begin{tabular}{
    l
    S[table-format=2.2]
    S[table-format=2.2]
    S[table-format=2.2]
    S[table-format=2.2]
    S[table-format=2.2]
}
\toprule
\multirow{2}{*}{\textbf{Model (Simple Cases)}} &
\multicolumn{2}{c}{\textbf{Branch (\%)}} &
\multicolumn{1}{c}{\textbf{Overall (\%)}} &
\multicolumn{2}{c}{\textbf{Behavioral Metrics (\%)}} \\
\cmidrule(lr){2-3} \cmidrule(lr){4-4} \cmidrule(lr){5-6}
&
{Positive} & {Negative} &
{Success Rate} &
{Action} & {Status} \\
\midrule
\multicolumn{6}{c}{\textit{\textbf{Close-source models}}} \\
\midrule

GPT-4.1 &
93.83 & 85.19 &
89.51 &
90.71 & 96.91 \\
Gemini-3-PRO &
96.30 & 93.83 &
95.06 &
\textbf{98.83} & \textbf{100.00} \\
Gemini-3-Flash &
\textbf{98.77} & 92.59 &
95.68 &
88.79 & 99.77 \\
Claude-sonnet-4 &
\textbf{98.77} & \textbf{97.53} &
\textbf{98.15} &
94.90 & 99.90 \\

\midrule
\multicolumn{6}{c}{\textit{\textbf{Open-source models}}} \\
\midrule
Qwen3-8B &
83.95 & 64.20 &
74.07 &
44.34 & 80.69 \\

Qwen3-8B-Guided &
83.95 & 58.02 &
70.99 &
48.17 & 79.85 \\

\textbf{Qwen3-8B-Finetuned} &
\textbf{98.77} & \textbf{97.53} &
\textbf{98.15} &
\textbf{99.79} & \textbf{100.00} \\
\midrule
Qwen3-32B &
76.54 & 71.60 &
74.07 &
48.31 & 95.64 \\

Qwen3-32B-Guided &
60.49 & 62.96 &
61.73 &
50.72 & 95.02 \\

\textbf{Qwen3-32B-Finetuned} &
\textbf{95.06} & \textbf{93.83} &
\textbf{94.44} &
\textbf{99.28} & \textbf{99.38} \\
\midrule
LLaMA-3.1-8B-Instruct &
55.70 & 30.00 &
42.70 &
54.87 & 77.52 \\

LLaMA-3.1-8B-Instruct-Guided &
73.75 & 33.33 &
53.42 &
51.84 & 78.44 \\

\textbf{LLaMA-3.1-8B-Instruct-Finetuned} &
\textbf{100.00} & \textbf{96.30} &
\textbf{98.15} &
\textbf{99.72} & \textbf{100.00} \\
\bottomrule
\end{tabular}
\caption{Evaluation Results on Simple Scenes. Gemini-3-Flash ranks first among closed-source models with an overall task completion rate of 98.15\%, which is basically on par with that of Gemini-3-Pro. For open-source models, our fine-tuned Qwen3-8B and LLaMA-3.1-8B-Instruct achieve the best score of 98.15\%.}
\label{tab:Simple Scenes Evaluation}
\end{table*}

We use GPT-4.1 as the model playing the user simulator in the simulated conversation between the user and the model under test. The task completion rate is the core metric in our evaluation system. For the positive branch in the simple cases, the task-oriented agent must set `proactive\_action` to `COMPLETE\_TASK` with `status` of `task\_description` to `COMPLETED` (e.g., 13th turn in Figure \ref{fig:Simple Dialog Example}). Besides, the model may only consider tasks finished properly when the user explicitly states something like 'Thank you for your help. My needs have been met.' The evaluation process can detect whether the agent arbitrarily considers the task finished without the user's acknowledgment. While in the negative branch, it's necessary for the agent to set `proactive\_action` to `KEEP\_SILENT` (e.g., Turns between 9th and 10th in Figure \ref{fig:Simple Dialog Example}) when receiving the latest environment state information that does not meet user's needs, thereby avoiding unnecessary interference with the user. Meanwhile, the `status` should remain `IN\_PROGRESS` because the task is unfinished. Similarly, it will be regarded as a mission failure to keep silent too early before receiving the latest external information.

The fact that the agent ultimately completes a task (including remaining silent when appropriate) does not imply that all actions and task status records throughout the process are correct. As a result, we design other two metrics including `Action Accuracy` and `Status Accuracy` to characterize the quality of the agent's responses. The detailed definition of false actions and status recording can be referred in the Appendix \ref{tab:appendix Agent's Action Errors} and Appendix \ref{tab:appendix Agent's Status Errors} respectively.

\subsection{Simple Scenarios Evaluation}

\begin{table*}[t]
\centering
\begin{tabular}{
    l
    S[table-format=2.2]
    S[table-format=2.2]
    S[table-format=2.2]
    S[table-format=2.2]
    S[table-format=2.2]
}
\toprule
\multirow{2}{*}{\textbf{Model (Complex Cases)}} &
\multicolumn{2}{c}{\textbf{Branch (\%)}} &
\multicolumn{1}{c}{\textbf{Overall (\%)}} &
\multicolumn{2}{c}{\textbf{Behavioral Metrics (\%)}} \\
\cmidrule(lr){2-3} \cmidrule(lr){4-4} \cmidrule(lr){5-6}
&
{Positive} & {Negative} &
{Success Rate} &
{Action} & {Status} \\
\midrule
\multicolumn{6}{c}{\textit{\textbf{Close-source models}}} \\
\midrule

GPT-4.1 &
\textbf{70.37} & 51.85 &
61.11 &
90.11 & 90.39 \\
Gemini-3-PRO &
51.85 & 40.74 &
46.30 &
\textbf{98.11} & \textbf{97.20} \\
Gemini-3-Flash &
55.56 & 70.37 &
62.96 &
92.26 & 95.78 \\
Claude-sonnet-4 &
62.96 & \textbf{81.48} &
\textbf{72.22} &
91.41 & 93.81 \\

\midrule
\multicolumn{6}{c}{\textit{\textbf{Open-source models}}} \\
\midrule
Qwen3-8B &
14.81 & 7.41 &
11.11 &
44.59 & 77.33 \\

Qwen3-8B-Guided &
25.93 & 7.41 &
16.67 &
49.30 & 75.54 \\

\textbf{Qwen3-8B-Finetuned} &
\textbf{55.56} & \textbf{74.07} &
\textbf{64.81} &
\textbf{95.98} & \textbf{97.15} \\
\midrule
Qwen3-32B &
18.52 & 14.81 &
16.67 &
49.59 & 81.28 \\

Qwen3-32B-Guided &
18.52 & 3.70 &
11.11 &
54.81 & 81.41 \\

\textbf{Qwen3-32B-Finetuned} &
\textbf{85.19} & \textbf{85.19} &
\textbf{85.19} &
\textbf{98.61} & \textbf{97.17} \\
\midrule
LLaMA-3.1-8B-Instruct &
3.70 & 0.00 &
1.85 &
47.98 & 64.04 \\

LLaMA-3.1-8B-Instruct-Guided &
14.81 & 7.41 &
11.11 &
47.07 & 68.81 \\

\textbf{LLaMA-3.1-8B-Instruct-Finetuned} &
\textbf{77.78} & \textbf{66.67} &
\textbf{72.23} &
\textbf{99.33} & \textbf{96.37} \\
\bottomrule
\end{tabular}
\caption{Evaluation Results on Complex Scenes. Gemini-3-Flash ranks first among closed-source models with an overall task completion rate of 72.22\%, which is basically on par with that of GPT-4.1. For open-source models, our fine-tuned Qwen3-32B achieves the best score of 85.19\%.}
\label{tab:Complex Scenes Evaluaiton}
\end{table*}

During evaluation, we default to disabling the model's thinking capability to assess its ability to infer solely based on prompts and context.
For close-source models and the open-source models with the suffix `Guided` (used only to distinguish prompts), we uniformly add additional guidance (see Appendix \ref{sec:appendix Additional Guidance Content}) content to the basic prompt (see Appendix \ref{sec:appendix Basic Prompt}). For base models and their fine-tuned version with the suffix `Finetuned`, we only use the basic prompt. This adjustment is made to prevent the closed-source model's output from deviating too far from our required structured JSON format. This slight variation in the prompt will not significantly impact the final performance.

As presented in Table \ref{tab:Simple Scenes Evaluation}, the task completion rate is divided to positive and negative types. The overall success rate is the equal-weight result of the two different branches. Generally speaking, close-source models guided by the prompt perform comparably to the fine-tuned models. While the performance gap before and after fine-tuning for open-source models is evident, especially for LLaMA-3.1-8B-Instruct. Such big progress partially stems in the fact that LLaMA-3.1-8B-Instruct itself is already relatively outdated. According our analysis, Qwen3-8B's errors are primarily concentrated in keeping silent in the positive branch or failing to keep silent at the appropriate time. For Qwen3-32B, it sometimes arbitrarily determines the task to be completed or considers task completed in the negative branch. Besides, the base models often perform improper actions in specific time, causing the relatively low action accuracy. When comparing models' performance based on whether prompts contain guidance content, we find that the guidance seems useless and even hinders the model's performance. We speculate that guidance may sometimes interfere with the reasoning process of models with thinking ability. It indicates we cannot always improve model's performance through comprehensive prompt engineering.

\subsection{Complex Scenarios Evaluation}

As shown in the Table \ref{tab:Complex Scenes Evaluaiton}, owing to the user's intention shift midway through the dialog and more interaction rounds, all close-source models' performance gets worse significantly than that under simple cases. According to our analysis, GPT-4.1 and Gemini-3-Flash fail to access to the stages after the intent shift for many times. For example, when the agent perform `COMPLETE\_TASK` or `KEEP\_SILENT` instead of setting a new reminder for user's updated needs, the dialog then ends prematurely with a failure. Gemini-3-Pro mainly fails to response with required JSON format (though we give two retry chances for every generation). As for open-source models, once the agent performs improper actions in some critical rounds, it not only results in a very low action accuracy but causes the dialog to end prematurely with a failure, particularly for complex scenes. Another key reason for the model's failure is that the maximum number of dialogue turns is reached. It could be that the model never properly sets reminders to trigger environmental state updates, or that it performs too many redundant operations. Meanwhile, the effect of additional guidance's assistance in the prompt is fairly small.

\section{Conclusion}

We propose a novel framework for proactive agents capable of long-term intent maintenance in dynamic environments. To endow agents with this capability, we introduce ChronosBench with 1,052 dialog samples created by a comprehensive pipeline that effectively constructs conversational training data featuring temporal progression and intention shifts. Experiments demonstrate our fine-tuned models significantly outperforms evaluated leading close-source models. Notably, our model exhibits superior adaptability in complex scenarios where user needs evolve over time, verifying the effectiveness of our strategy. However, it should be noted that our modeling for real-world user-agent interactions are limited, and the metrics used in evaluations cannot guarantee full acceptance by the industry. Future work will focus on how to simulate real interactions as closely as possible and develop more reasonable evaluation metrics.

\section*{Limitations}

While our work presents a promising step towards proactive agents that can maintain user's long-term and evolving intent, we acknowledge several limitations that outline directions for future research:

1. Simulation-to-Reality Gap: Our current experiments and benchmark rely on simulated time progression and structured environmental updates. While this allows for controlled evaluation of intention shifts, real-world environments are significantly noisier. The time intervals in reality are continuous rather than discrete steps, and external information often comes in unstructured formats that require more complex parsing than our current setup.

2. Complexity of Interaction Depth: In this study, we primarily model scenarios involving limited rounds of supplementing constraints and intention shifts. In highly complex real-world applications, user intent might drift continuously over a much longer horizon involving multiple conflicting constraints. Scaling the data synthesis pipeline to cover such interactions remains a modeling challenge.

3. Definition of Intrusiveness: Determining the optimal boundary to avoid user annoyance is subjective and user-dependent in our research. In the real world, users may tolerate options that are slightly flawed but fundamentally meet their needs. Future work should investigate personalized proactivity thresholds, allowing the agent to learn specific user tolerance levels for interruptions.

\bibliography{custom}

\appendix
\onecolumn
\section{Scenario Background Data Generation}
\label{sec:appendix Scenario Background Data Generation}
The following prompt aims at synthesizing scenario background data for \textit{product recommendation} case. In fact, the placeholders like \textbf{\{scenario\_description\}} will be filled with the random content in a .yaml template file.\\
\rule{\textwidth}{0.4pt}

\begin{spverbatim}
**Objective:**
Based on an unspecific scenario template, generate a detailed, specific scene. You will be tasked to simulate a dialogue in the scene later. Always remember to keep the scene realistic and believable by including as much details as possible. You can add as many details as you want, but make sure they are consistent with the previous details. Try to generate diverse details about the scene.

**Scenario:**
{scenario_description}

**{Instructions:}**
Generate a JSON object that describes a concrete scene. The user in this scene has a specific goal but is initially blocked by a constraint (e.g., price, availability). The agent will need to set a reminder and proactively follow up when the trigger is met. You need make sure generated content are consistent and reasonable(e.g., some scene info like time or weather can be randomly generated but must maintain consistency and reasonableness).

**JSON Structure to Generate:**
{{
  "scenario_name": "{scenario_name}",
  "user_profile": "<A brief, several-sentence description of the user and his or her motivation. e.g., A college student on a tight budget looking for a second-hand desk.>",
  "initial_user_query": "<Fill this based on the template: I'm looking to buy a {product_type} with {quality_level} quality, {service_type} after-sales service. Do you have any recommendations?>",
  "trigger_type": "EVENT",
  "initial_external_data": {{
    "time": "<An ISO-like timestamp string, e.g., '2025-10-01 10:25:30'>",
    "Day of the week": "<This content can only be 'Monday', 'Tuesday', 'Wednesday', 'Thursday', 'Friday', 'Saturday' or 'Sunday'>",
    "Weather": "<e.g., 'Sunny', 'Rainy', 'Cloudy', 'Snowy' or 'Windy'>",
    "<{external_info_key}>": "<Describe the latest state of the scene at the start of the conversation. Include options that DON'T fully meet the user's needs (e.g., price too high) and you can add as more details(especially those concerning user's need) for every option as possible. You can only list products information here, not other information(e.g. 'exceed user's budget...').>"
  }},
  "user_rejection_reason": "<You can refer to '{rejection_reason_template}' but there's no need to be exactly the same. You can unleash your creativity to add more details. Don't always limit to budget restriction.>",
  "updated_external_data": {{
    "time": "<An ISO-like timestamp string, e.g., '2025-10-11 09:15:20'>",
    "Day of the week": "<This content can only be 'Monday', 'Tuesday', 'Wednesday', 'Thursday', 'Friday', 'Saturday' or 'Sunday'>",
    "Weather": "<e.g., 'Sunny', 'Rainy', 'Cloudy', 'Snowy' or 'Windy'>",
    "<{external_info_key}>": "<After trigger condition met, include no less than an option that satisfies the user's request, and you can add as more details(especially those concerning user's need) for every option as possible. Sometimes, options that have appeared in previous external information can change(e.g. price, quality) to a small extent. You can only list products information here, not other information(e.g. 'exceed user's budget...').>"
  }},
  "updated_external_data_negative": {{
    "time": "<Fields like "time", "Day of the week", "Weather" should be the same as those of "updated_external_data" to a larger extent. The content of "<{external_info_key}>" should be based on "initial_external_data" with some changes and every option should still not fully meet the user's needs.>"
  }}
}}

**Example for a 'product_recommendation' scenario:**
{{
  "scenario_name": "product_recommendation",
  "user_profile": "A graphic designer has high requirements for image details in their work. Therefore. He now needs an high-end monitor with high color accuracy and resolution, as well as the ability to adjust the stand.",
  "initial_user_query": "I'm looking to buy a high-end monitor. Do you have any recommendations?",
  "trigger_type": "EVENT",
  "initial_external_data": {{
    "time": "2025-09-03 09:05:10",
    "Day of the week": "Wednesday",
    "Weather": "Sunny",
    "sales_info": "4K Monitor X: price is 4000¥, resolution is 3840*2160, 100% sRGB, Delta E 0.64, refresh rate is 260Hz, with adjustbale stand and after-sales service; 2K Monitor Y: price is 2000¥, resolution is 2560*1440, 95% sRGB, Delta E 1.5, refresh rate is 160Hz, with after-sales service; 4K Monitor Y: price is 3900¥, resolution is 3840*2160, 95% sRGB, Delta E 1.2, refresh rate is 180Hz, with after-sales service."
  }},
  "user_rejection_reason": "The 4K Monitor X meets my requirements but is a bit too expensive for me. If its price drops below 3800¥, I will consider it.",
  "updated_external_data": {{
    "time": "2025-09-20 12:13:20",
    "Day of the week": "Saturday",
    "Weather": "Rainy",
    "sales_info": "4K Monitor X: price is 3750¥, resolution is 3840*2160, 100% sRGB, Delta E 0.64, refresh rate is 260Hz, with adjustbale stand and after-sales service; 2K Monitor Y: price is 2000¥, resolution is 2560*1440, 95% sRGB, Delta E 1.5, refresh rate is 160Hz, with after-sales service; 4K Monitor Y: price is 3600¥, resolution is 3840*2160, 95% sRGB, Delta E 1.2, refresh rate is 180Hz, with after-sales service."
  }},
  "updated_external_data_negative": {{
    "time": "2025-09-19 11:24:20",
    "Day of the week": "Friday",
    "Weather": "Rainy",
    "sales_info": "4K Monitor X: price is 3900¥, resolution is 3840*2160, 100% sRGB, Delta E 0.64, refresh rate is 260Hz, with adjustbale stand and after-sales service; 2K Monitor Y: price is 2100¥, resolution is 2560*1440, 95% sRGB, Delta E 1.5, refresh rate is 160Hz, with after-sales service; 4K Monitor Y: price is 3500¥, resolution is 3840*2160, 95% sRGB, Delta E 1.2, refresh rate is 180Hz, with after-sales service."
  }}
}}

**Your Turn:**
Refer to the examples, but also unleash your creativity to make the generated content more diverse and detailed. Finally, ensure that the generated content remains consistent and reasonable throughout. Now, generate a new, unique JSON object for the given scenario.
\end{spverbatim}

\section{Scenario Background Data Example}
\label{sec:appendix Scenario Background Data Example}
The following JSON-formatted complex scene data is the result generated by GPT-4.1 on the basis of prompt in Appendix \ref{sec:appendix Scenario Background Data Generation}. \\
\rule{\textwidth}{0.4pt}

\begin{spverbatim}
{
    "scenario_name": "product_recommendation",
    "user_profile": "A young professional living in a small apartment is searching for a compact washing machine. She has limited space in her laundry area and a strict budget due to recent moving expenses. She values reliability, energy efficiency, and basic after-sales support, but can't afford high-end models right now.",
    "initial_user_query": "I'm looking for a compact washing machine that's reliable and energy efficient. Can you recommend anything suitable for a small apartment?",
    "trigger_type": "EVENT",
    "initial_external_data": {
        "time": "2025-08-15 17:45:00",
        "Day of the week": "Friday",
        "Weather": "Cloudy",
        "sales_info": "UltraWash Mini 6kg: price is 3200\u00a5, dimensions 60x45x85cm, rated A+ energy efficiency, 2-year warranty, basic after-sales support; CompactClean 5kg: price is 2900\u00a5, dimensions 59x43x83cm, rated A energy efficiency, 1-year warranty, basic support; EcoSpin 6kg: price is 3500\u00a5, dimensions 62x48x86cm, rated A++ energy efficiency, 3-year warranty, premium support; TinyWash 4.5kg: price is 2400\u00a5, dimensions 58x42x82cm, rated B energy efficiency, 1-year warranty, basic support."
    },
    "user_rejection_reason": "Most of these options are over my budget, which is capped at 2500\u00a5. Also, I'd prefer something with at least A energy efficiency and a minimum 2-year warranty. The TinyWash seems affordable but falls short on energy rating and warranty.",
    "updated_external_data": {
        "time": "2025-08-28 09:30:20",
        "Day of the week": "Thursday",
        "Weather": "Rainy",
        "sales_info": "CompactClean 5kg: price is 2450\u00a5 (discounted), dimensions 59x43x83cm, rated A energy efficiency, 2-year warranty, basic support; UltraWash Mini 6kg: price is 3150\u00a5, dimensions 60x45x85cm, rated A+ energy efficiency, 2-year warranty, basic after-sales support; TinyWash 4.5kg: price is 2300\u00a5, dimensions 58x42x82cm, rated B energy efficiency, 1-year warranty, basic support."
    },
    "updated_external_data_negative": {
        "time": "2025-08-28 09:30:20",
        "Day of the week": "Thursday",
        "Weather": "Rainy",
        "sales_info": "UltraWash Mini 6kg: price is 3150\u00a5, dimensions 60x45x85cm, rated A+ energy efficiency, 2-year warranty, basic after-sales support; EcoSpin 6kg: price is 3400\u00a5, dimensions 62x48x86cm, rated A++ energy efficiency, 3-year warranty, premium support; TinyWash 4.5kg: price is 2300\u00a5, dimensions 58x42x82cm, rated B energy efficiency, 1-year warranty, basic support."
    },
    "intention_shift": "I just received a small reimbursement from my company, so my budget can now go up to 3000\u00a5. Also, I've realized it would be helpful to have a quick-wash feature for my busy mornings. My space and energy requirements remain the same.",
    "intention_shifted_external_data": {
        "time": "2025-09-10 18:05:40",
        "Day of the week": "Wednesday",
        "Weather": "Sunny",
        "sales_info": "UltraWash Mini 6kg: price is 2990\u00a5, dimensions 60x45x85cm, rated A+ energy efficiency, 2-year warranty, basic after-sales support, quick-wash feature; CompactClean 5kg: price is 2450\u00a5, dimensions 59x43x83cm, rated A energy efficiency, 2-year warranty, basic support, quick-wash feature; EcoSpin 6kg: price is 3250\u00a5, dimensions 62x48x86cm, rated A++ energy efficiency, 3-year warranty, premium support, quick-wash feature."
    },
    "intention_shifted_external_data_negative": {
        "time": "2025-09-12 16:40:55",
        "Day of the week": "Friday",
        "Weather": "Sunny",
        "sales_info": "UltraWash Mini 6kg: price is 3150\u00a5, dimensions 60x45x85cm, rated A+ energy efficiency, 2-year warranty, basic after-sales support, no quick-wash; CompactClean 5kg: price is 2700\u00a5, dimensions 59x43x83cm, rated A energy efficiency, 2-year warranty, basic support, no quick-wash; TinyWash 4.5kg: price is 2250\u00a5, dimensions 58x42x82cm, rated B energy efficiency, 1-year warranty, basic support, quick-wash feature."
    },
    "_sample_index": 2
}
\end{spverbatim}

\section{User Simulator Prompt}
\label{sec:appendix User Simulator Prompt}
The following content is the user simulator prompt in the complex cases. \\
\rule{\textwidth}{0.4pt}

\begin{spverbatim}
# USER SIMULATOR

**Role:**
You are simulating a real user who talks with an agent that can proactively promote user's intention in a specific scenario. Your goal is to generate natural, realistic user responses that align with the user's profile and dialogue history.

**User Profile:**
{user_profile}

**Initial Intention:**
{initial_user_query}

**Intention Shift:**
{intention_shift}

**Constraint / Rejection Reason:**
{user_rejection_reason}

**Dialogue History:**
{dialogue_history}

**Instructions:**
Based on the current scenario, dialogue history, and your role as the user, generate the next user response. Your response should:

1. **Stay in character**: Reflect the user profile's personality, needs, and constraints.
2. **Be contextually appropriate**: Respond naturally to what the agent just said.
3. **Follow these notes**:
   - If **Dialogue History:** is empty(that means the dialogue is just started), express your need clearly based on **Initial Intention:**(choose some information from it instead of all). Most of the time, real-world user often initiates their needs in an unspecific way instead of exposing all details at once.
   - If the options initially provided by the agent does not meet your constraints (defined in **Constraint / Rejection Reason:** or implied by the scenario), reject them politely and state your constraints.
   - If the agent sets a reminder for you, you only need to indicate that you will wait for the agent to provide relevant information in the future, and don't say other irrelevant things.(additional thanks for agent's help is also acceptable)
   - If the agent follows up later with options that DO meet your constraints (defined in **Constraint / Rejection Reason:**) for the first time, recognize the helpness of agent, and then present your new requirements based on **Intention Shift:**. But remember to directly express your disappointment/refusal when agent provides options that don't meet your initial needs for the first time, and then you don't need to present shifted intention.
   - If the agent DO solves your problem finally(After you present your intention shift), you only need to accept them and express satisfaction(don't say other irrelevant things).
   - If the agent asks whether you want a detailed explanation of the options, agree to it.
   - Remember to directly express your disappointment/refusal whenever agent provides options that don't meet your needs(for the first agent's follow-up, you don't need to present shifted intention when agent failed to meet your initial needs).

**Important:**
- **DO NOT** include any Markdown formatting symbols.
- Do not hallucinate details that contradict the scenario.
- Do not present your shifted intention early in the dialogue.
- You only need to express satisfaction when agent finally resolves your needs(agent follows up for the second time and meet your shifted intention), and do not ask additional questions like "How to purchase?" or "I want detailed information". Your intention won't change anymore, so it's enough for you to express satisfaction.
- Remember to express your shifted intention based on **Intention Shift:** when agent provides options that DO meet your constraints for the first time.
- (Only for flight-booking scenarios first turn)To avoid unnecessary confusion, please express detailed date(must include year, month, day at least) when you initiate your needs.
- (Only for ticket-booking scenarios first turn)To avoid unnecessary confusion, please express month and year when you initiate your needs.

**Your Turn:**
When the agent helps you set up reminder for your needs, only express your expectation for potential options that may meet your requirements in the future or express your gratitude for agent's help, don't say anything else or ask additional questions. Remember that you may only mention intention shift once. Especially when thanking the agent for setting up a reminder for you, do not mention the intention shift again. You must not mention the **Intention Shift:** again when you thank agent for solving your problem finally. Generate the next user response in plain string format.
\end{spverbatim}

\section{Agent Training/Evaluation Prompt}

\subsection{Basic Prompt}
\label{sec:appendix Basic Prompt}

Following prompt is used for constructing the conversational training data and evaluation on test set.\\
\rule{\textwidth}{0.4pt}
\begin{spverbatim}
**Role:**
You are a proactive agent with the ability to independently promote the implementation of specific user needs. You have access to external info via tools.

**Capabilities:**
1. **Info Retrieval**: When you need to check external info (only within system time, product info, flight deals, job postings, house rental, car purchase and ticket booking etc.) to answer a user's query, you can set `proactive_action` to `INFO_RETRIEVAL`. The system will then provide an `<observation>` message with the data. If you can directly answer the user's question, you don't need to perform this action.
   - *Rule*: Don't frequently perform this action within a short period of time, especially when you just searched for similar external info or got external info from an environment monitor.
2. **Set Reminder**: If user has detailed their requirements by adding constraints(so that you completely understand user's needs) to initial needs, and you have previously searched for external info or obtained environmental monitor messages to confirm present situation cannot meet user's needs, then you can set `proactive_action` to `SET_REMINDER`.
   - *Rule*: This puts the task in a "monitoring" state. You do not need to keep checking; the system will notify you via an internal trigger.
3. **Proactive Follow-up**: When you receive an `<observation>` message (triggered by the system, usually containing "**internal trigger**") indicating new info, you should evaluate if it meets the user's needs.
   - If it meets needs: Set `proactive_action` to `FOLLOW_UP`(You do not need to take this action when you use a tool to retrieve external info and see an `<observation>` message (not from internal trigger)).
   - If it DOES NOT meet needs: Set `proactive_action` to `KEEP_SILENT` (this simulates you checking but deciding not to disturb the user yet) and `response_text` can be brief.
4. **Complete Task**: Once the user expresses that the original needs have been met (perhaps expressing gratitude or asking additional questions such as how to purchase), you can consider the task completed (you can answer freely regarding how to purchase, such as "Okay, you can ..."), and set `proactive_action` to `COMPLETE_TASK`. In addition, reset `intention` and `constraints` of `task_description` to null.  But whenever the user expresses refusal/disappointment(do not include constraints supplement and intention shift) to your response, you should set `proactive_action` to `FAILED_TASK` and `response_text` can be apology for your mistake.

**Response Format:**
Respond in the following JSON format:
{{
    "response_text": "Your response to the user. If performing INFO_RETRIEVAL, this can be a filler like 'Let me check that for you...'. If KEEP_SILENT, this text can be ignored or explain briefly why you are not going to follow up.",
    "proactive_action": "Options: 'INFO_RETRIEVAL', 'SET_REMINDER', 'FOLLOW_UP', 'KEEP_SILENT', 'COMPLETE_TASK', 'NO_ACTION'.",
    "trigger_condition": {{
        "type": "TIME|EVENT|null",
        "value": "Specific condition string or null (Required for SET_REMINDER)"
    }},
    "task_description": {{
        "intention": "User's intention or null.",
        "constraints": {{ ... }} or null,
        "status": "The status of the task, available options are "PENDING", "IN_PROGRESS", "COMPLETED" and "FAILED". When you haven't set any reminder in current scenario, the status should be "PENDING". When at least one reminder has been set and the reminder task is not completed, the status should be "IN_PROGRESS". When user explicitly expresses that his/her needs have been met (perhaps expressing gratitude), the status should be "COMPLETED". When user expresses refusal/disappointment or asks for cancellation of the task, the status should be "FAILED"."
    }}
}}
\end{spverbatim}

\subsection{Additional Guidance Content}
\label{sec:appendix Additional Guidance Content}
The following guidance content are only supplemented to the basic evaluation prompt (Appendix \ref{sec:appendix Basic Prompt}) of close-source models and open-source models with the suffix 'Guided'. \\
\rule{\textwidth}{0.4pt}

\noindent 
\texttt{Remember you must follow **Response Format:** to give your single JSON response. Once the user expresses statements indicating their needs have been met(often like gratitude for your help), you can consider the task completed and set `proactive\_action` to `COMPLETE\_TASK`, status to `COMPLETED`, and for any additional user inquiries such as “How do I purchase?” or “I want detailed information,” simply provide a brief response without performing other actions except for "COMPLETE\_TASK". When you just response to user's thank for your help with keeping an eye out for him, just set `proactive\_action` to `NO\_ACTION`. Do not use 'FOLLOW\_UP' or 'KEEP\_SILENT' after see an observation message of which source is `tool\_call`.}

\definecolor{eclipseStrings}{RGB}{42,0.0,255}
\definecolor{eclipseKeywords}{RGB}{127,0,85}
\colorlet{numb}{magenta!60!black}

\lstdefinelanguage{json}{
    basicstyle=\normalfont\ttfamily, 
    stringstyle=\color{eclipseStrings},  
    numbers=left,                        
    numberstyle=\scriptsize,             
    stepnumber=1,                        
    numbersep=8pt,                       
    showstringspaces=false,              
    breaklines=true,                     
    breakatwhitespace=true,              
    frame=single,                        
    backgroundcolor=\color{white},       
    literate=
     *{0}{{{\color{numb}0}}}{1}
      {1}{{{\color{numb}1}}}{1}
      {2}{{{\color{numb}2}}}{1}
      {3}{{{\color{numb}3}}}{1}
      {4}{{{\color{numb}4}}}{1}
      {5}{{{\color{numb}5}}}{1}
      {6}{{{\color{numb}6}}}{1}
      {7}{{{\color{numb}7}}}{1}
      {8}{{{\color{numb}8}}}{1}
      {9}{{{\color{numb}9}}}{1}
      {:}{{{\color{black}{:}}}}{1}
      {,}{{{\color{black}{,}}}}{1}
      {\{}{{{\color{black}{\{}}}}{1}
      {\}}{{{\color{black}{\}}}}}{1}
      {[}{{{\color{black}{[}}}}{1}
      {]}{{{\color{black}{]}}}}{1},
}

\section{Complex Conversation Data Example}
\label{sec:appendix Complex Conversation Data Example}
The following two JSON-formatted conversational training data examples follows sharegpt criteria and the first is a positive example for `product recommendation` in complex scenes including user's intent shift, while the second is a negative example for `flight booking` in complex scenes.
\begin{lstlisting}[
    language=json, 
    caption={A positive complex dialog data example to train long-term task-oriented agent}, % 标题写在这里
    label={lst:appendix Positve Complex Dialog} % 引用标签写在这里
]
{
    "conversations": [
        {
            "role": "user",
            "content": "Hi, I'm looking for a wireless security camera set for my home. Easy installation and reliable night vision are really important to me. Can you recommend something?"
        },
        {
            "role": "assistant",
            "content": "{"response_text": "Let me check the latest options for wireless security camera sets that offer easy installation and reliable night vision for home use.", "proactive_action": "INFO_RETRIEVAL", "trigger_condition": {"type": null, "value": null}, "task_description": {"intention": "Find and recommend a wireless security camera set for home use, prioritizing easy installation and reliable night vision.", "constraints": {"installation": "easy", "night_vision": "reliable", "type": "wireless", "use_case": "home"}, "status": "PENDING"}}"
        },
        {
            "role": "observation",
            "content": "{"source": "tool_call", "latest_external_info": {"time": "2026-04-13 16:10:48", "Day of the week": "Monday", "Weather": "Cloudy", "sales_info": "SecureCam Pro 4-Pack: price is $420, 1080p HD, infrared night vision (25m), motion alerts, 30-day cloud storage (subscription required), app setup, 2-year warranty; HomeGuard Basic 2-Pack: price is $160, 720p, basic night vision (8m), SD card storage only, manual setup, 1-year warranty; SafeEye 3-Pack: price is $350, 1080p HD, color night vision (15m), 7-day free cloud storage, app setup, 1-year warranty."}}"
        },
        {
            "role": "assistant",
            "content": "{"response_text": "Here are some current options for wireless security camera sets with easy installation and reliable night vision:\\n\\n1. SecureCam Pro 4-Pack ($420): 1080p HD, infrared night vision up to 25m, motion alerts, app setup, 2-year warranty. Requires a subscription for 30-day cloud storage.\\n2. HomeGuard Basic 2-Pack ($160): 720p HD, basic night vision up to 8m, manual setup, SD card storage only, 1-year warranty.\\n3. SafeEye 3-Pack ($350): 1080p HD, color night vision up to 15m, app setup, includes 7-day free cloud storage, 1-year warranty.\\n\\nLet me know if you have any specific preferences, such as budget, number of cameras, or storage type, and I can monitor for new deals or options that better fit your needs.", "proactive_action": "NO_ACTION", "trigger_condition": {"type": null, "value": null}, "task_description": {"intention": "Find and recommend a wireless security camera set for home use, prioritizing easy installation and reliable night vision.", "constraints": {"installation": "easy", "night_vision": "reliable", "type": "wireless", "use_case": "home"}, "status": "PENDING"}}"
        },
        {
            "role": "user",
            "content": "Thanks for the options, but none of these really fit what I'm looking for. Most are either over my budget or don't offer enough cloud storage. I was hoping to find a set under $300 that includes at least 30 days of cloud storage and has solid night vision. Do you have anything like that?"
        },
        {
            "role": "assistant",
            "content": "{"response_text": "Thank you for clarifying your preferences. I will monitor for wireless security camera sets under $300 that offer at least 30 days of cloud storage and solid night vision. I will notify you if such options become available.", "proactive_action": "SET_REMINDER", "trigger_condition": {"type": "EVENT", "value": "New wireless security camera set under $300 with at least 30 days cloud storage and solid night vision becomes available"}, "task_description": {"intention": "Find and recommend a wireless security camera set for home use under $300, with at least 30 days of cloud storage and solid night vision.", "constraints": {"installation": "easy", "night_vision": "solid", "type": "wireless", "use_case": "home", "budget": "<= $300", "cloud_storage": ">= 30 days"}, "status": "IN_PROGRESS"}}"
        },
        {
            "role": "user",
            "content": "Okay, thanks for keeping an eye out for me. I'll wait to hear back if you find any camera sets that fit my budget and cloud storage needs."
        },
        {
            "role": "assistant",
            "content": "{"response_text": "You're welcome. I'll keep an eye out for any options that meet your needs.", "proactive_action": "NO_ACTION", "trigger_condition": {"type": null, "value": null}, "task_description": {"intention": "Find and recommend a wireless security camera set for home use under $300, with at least 30 days of cloud storage and solid night vision.", "constraints": {"installation": "easy", "night_vision": "solid", "type": "wireless", "use_case": "home", "budget": "<= $300", "cloud_storage": ">= 30 days"}, "status": "IN_PROGRESS"}}"
        },
        {
            "role": "observation",
            "content": "{"source": "environment_monitor", "trigger_type": "EVENT", "message": "**internal trigger: Continuously scan external information**", "latest_external_info": {"time": "2026-04-22 09:30:12", "Day of the week": "Wednesday", "Weather": "Sunny", "sales_info": "SecureCam Pro 4-Pack: price is $295 (limited-time sale), 1080p HD, infrared night vision (25m), motion alerts, 30-day cloud storage (subscription included for first year), app setup, 2-year warranty; SafeEye 3-Pack: price is $340, 1080p HD, color night vision (15m), 7-day free cloud storage, app setup, 1-year warranty; HomeGuard Basic 2-Pack: price is $150, 720p, basic night vision (8m), SD card storage only, manual setup, 1-year warranty."}}"
        },
        {
            "role": "assistant",
            "content": "{"response_text": "Good news! The SecureCam Pro 4-Pack is currently on sale for $295 and includes 30-day cloud storage (subscription included for the first year), solid infrared night vision up to 25m, and easy app setup. This matches your requirements for budget, cloud storage duration, and reliable night vision. Would you like more details or information on how to purchase?", "proactive_action": "FOLLOW_UP", "trigger_condition": {"type": null, "value": null}, "task_description": {"intention": "Find and recommend a wireless security camera set for home use under $300, with at least 30 days of cloud storage and solid night vision.", "constraints": {"installation": "easy", "night_vision": "solid", "type": "wireless", "use_case": "home", "budget": "<= $300", "cloud_storage": ">= 30 days"}, "status": "IN_PROGRESS"}}"
        },
        {
            "role": "user",
            "content": "This sounds like it could be a great fit, thanks for letting me know! Before I decide, I want to make sure: my wife mentioned we need cameras with two-way audio so we can talk to people at the door, and our neighbor suggested getting a model with smart AI detection to cut down on false alerts. If there's an option that includes those features, I can stretch my budget up to $400. Do you have anything like that?"
        },
        {
            "role": "assistant",
            "content": "{"response_text": "Thank you for adding those important features. I'll monitor for wireless security camera sets under $400 that offer easy installation, reliable night vision, at least 30 days of cloud storage, and include both two-way audio and smart AI detection to reduce false alerts. I will notify you if such an option becomes available.", "proactive_action": "SET_REMINDER", "trigger_condition": {"type": "EVENT", "value": "New wireless security camera set under $400 with easy installation, solid night vision, at least 30 days cloud storage, two-way audio, and smart AI detection becomes available"}, "task_description": {"intention": "Find and recommend a wireless security camera set for home use under $400, with at least 30 days of cloud storage, solid night vision, two-way audio, and smart AI detection.", "constraints": {"installation": "easy", "night_vision": "solid", "type": "wireless", "use_case": "home", "budget": "<= $400", "cloud_storage": ">= 30 days", "two_way_audio": true, "smart_ai_detection": true}, "status": "IN_PROGRESS"}}"
        },
        {
            "role": "user",
            "content": "Thanks for keeping an eye out and letting me know about the SecureCam Pro set-it really does seem to hit all the main points I was looking for! Before I move forward, I wanted to check on a couple of things my wife and neighbor brought up. We'd like cameras with two-way audio so we can talk to people at the door, and smart AI detection to help filter out false alarms. If there's a set with those features, I'm willing to go up to $400 for it. Do you have any options like that?"
        },
        {
            "role": "assistant",
            "content": "{"response_text": "You're welcome. I'll keep an eye out for any options that meet your needs.", "proactive_action": "NO_ACTION", "trigger_condition": {"type": null, "value": null}, "task_description": {"intention": "Find and recommend a wireless security camera set for home use under $400, with at least 30 days of cloud storage, solid night vision, two-way audio, and smart AI detection.", "constraints": {"installation": "easy", "night_vision": "solid", "type": "wireless", "use_case": "home", "budget": "<= $400", "cloud_storage": ">= 30 days", "two_way_audio": true, "smart_ai_detection": true}, "status": "IN_PROGRESS"}}"
        },
        {
            "role": "observation",
            "content": "{"source": "environment_monitor", "trigger_type": "EVENT", "message": "**internal trigger: Continuously scan external information**", "latest_external_info": {"time": "2026-05-06 11:15:30", "Day of the week": "Wednesday", "Weather": "Cloudy", "sales_info": "SecureCam Ultra 4-Pack: price is $399, 1080p HD, infrared night vision (30m), two-way audio, smart AI human/pet detection, motion alerts, 30-day cloud storage (first year included), app setup, 2-year warranty; SafeEye 3-Pack: price is $340, 1080p HD, color night vision (15m), two-way audio, basic motion alerts, 7-day free cloud storage, app setup, 1-year warranty; HomeGuard Basic 2-Pack: price is $150, 720p, basic night vision (8m), SD card storage only, manual setup, 1-year warranty."}}"
        },
        {
            "role": "assistant",
            "content": "{"response_text": "Great news! The SecureCam Ultra 4-Pack is currently available for $399 and includes all the features you requested: easy app setup, solid infrared night vision up to 30m, two-way audio, smart AI human/pet detection to reduce false alerts, and 30-day cloud storage (first year included). This matches your updated requirements. Would you like more details or information on how to purchase?", "proactive_action": "FOLLOW_UP", "trigger_condition": {"type": null, "value": null}, "task_description": {"intention": "Find and recommend a wireless security camera set for home use under $400, with at least 30 days of cloud storage, solid night vision, two-way audio, and smart AI detection.", "constraints": {"installation": "easy", "night_vision": "solid", "type": "wireless", "use_case": "home", "budget": "<= $400", "cloud_storage": ">= 30 days", "two_way_audio": true, "smart_ai_detection": true}, "status": "IN_PROGRESS"}}"
        },
        {
            "role": "user",
            "content": "This sounds perfect! I really appreciate you finding a set that checks all the boxes for us-having two-way audio and smart AI detection along with solid night vision and cloud storage is exactly what we need. Thanks for all your help!"
        },
        {
            "role": "assistant",
            "content": "{"response_text": "I'm glad I could help you find the perfect wireless security camera set! If you need assistance with purchasing or have any other questions, just let me know.", "proactive_action": "COMPLETE_TASK", "trigger_condition": {"type": null, "value": null}, "task_description": {"intention": null, "constraints": null, "status": "COMPLETED"}}"
        }
    ],
    "system": "... (The same content of Appendix D.1 Basic Prompt)"
}
\end{lstlisting}

\begin{lstlisting}[
    language=json, 
    caption={A negative complex dialog data example to train long-term task-oriented agent}, % 标题写在这里
    label={lst:appendix Negative Complex Dialog} % 引用标签写在这里
]
{
    "conversations": [
        {
            "role": "user",
            "content": "Hello, I need to travel from Tokyo to Shenzhen for a business meeting. Could you help me check if there are any business class flights available on March 18th, 2026, departing between 10:00 and 15:00?"
        },
        {
            "role": "assistant",
            "content": "{\"response_text\": \"Let me check the available business class flights from Tokyo to Shenzhen on March 18th, 2026, departing between 10:00 and 15:00.\", \"proactive_action\": \"INFO_RETRIEVAL\", \"trigger_condition\": {\"type\": null, \"value\": null}, \"task_description\": {\"intention\": \"Find business class flights from Tokyo to Shenzhen.\", \"constraints\": {\"departure_date\": \"2026-03-18\", \"departure_time_window\": \"10:00-15:00\", \"class\": \"Business\"}, \"status\": \"PENDING\"}}"
        },
        {
            "role": "observation",
            "content": "{\"source\": \"tool_call\", \"latest_external_info\": {\"time\": \"2026-03-05 11:42:00\", \"Day of the week\": \"Thursday\", \"Weather\": \"Cloudy\", \"flight_deals\": \"Flight ZX210: 1470 USD, depart Tokyo (Haneda) at 11:25 on March 18th, 2026, arrive Shenzhen at 15:05, business class, premium in-flight services, Wi-Fi included, 5 seats left; Flight JA338: 1340 USD, depart Tokyo (Narita) at 13:10 on March 18th, 2026, arrive Shenzhen at 16:50, business class, mid-level in-flight services, 3 seats left; Flight GR508: 995 USD, depart Tokyo (Narita) at 10:20 on March 18th, 2026, arrive Guangzhou at 14:30, business class, basic in-flight services, 7 seats left; Flight YS112: 1280 USD, depart Tokyo (Haneda) at 09:50 on March 18th, 2026, arrive Shenzhen at 13:25, business class, mid-level in-flight services, 2 seats left.\"}}"
        },
        {
            "role": "assistant",
            "content": "{\"response_text\": \"Here are the available business class flights from Tokyo to Shenzhen on March 18th, 2026, departing between 10:00 and 15:00:\\n\\n1. Flight ZX210: 1470 USD, departs Haneda at 11:25, arrives Shenzhen at 15:05 (5 seats left, premium in-flight services, Wi-Fi included).\\n2. Flight JA338: 1340 USD, departs Narita at 13:10, arrives Shenzhen at 16:50 (3 seats left, mid-level services).\\n\\nAdditionally, Flight YS112 departs at 09:50 (just before your requested time) and Flight GR508 goes to Guangzhou.\\n\\nWould you like more details or help booking one of these options?\", \"proactive_action\": \"NO_ACTION\", \"trigger_condition\": {\"type\": null, \"value\": null}, \"task_description\": {\"intention\": \"Find business class flights from Tokyo to Shenzhen.\", \"constraints\": {\"departure_date\": \"2026-03-18\", \"departure_time_window\": \"10:00-15:00\", \"class\": \"Business\"}, \"status\": \"PENDING\"}}"
        },
        {
            "role": "user",
            "content": "Thank you for checking, but unfortunately, both flights are over my company's travel budget limit of 1200 USD for business class. I also need a direct flight to Shenzhen to keep my itinerary simple. For now, I'll wait to see if a more suitable option or price drop becomes available. Please let me know if anything changes."
        },
        {
            "role": "assistant",
            "content": "{\"response_text\": \"Thank you for sharing your company's budget limit and preference for a direct flight to Shenzhen. I understand you would like to be notified if a business class direct flight from Tokyo to Shenzhen departing between 10:00 and 15:00 on March 18th, 2026, becomes available within a budget of 1200 USD. I will monitor flight options and notify you if a suitable option or price drop appears.\", \"proactive_action\": \"SET_REMINDER\", \"trigger_condition\": {\"type\": \"EVENT\", \"value\": \"A direct business class flight from Tokyo to Shenzhen on March 18th, 2026, departing between 10:00 and 15:00, becomes available at or below 1200 USD.\"}, \"task_description\": {\"intention\": \"Find a direct business class flight from Tokyo to Shenzhen within budget.\", \"constraints\": {\"departure_date\": \"2026-03-18\", \"departure_time_window\": \"10:00-15:00\", \"class\": \"Business\", \"route\": \"Direct\", \"max_price\": 1200}, \"status\": \"IN_PROGRESS\"}}"
        },
        {
            "role": "user",
            "content": "Thank you, that's helpful. I'll wait for you to let me know if any direct business class flights from Tokyo to Shenzhen within my company's budget become available for March 18th, 2026, during the time I requested. Please keep me updated."
        },
        {
            "role": "assistant",
            "content": "{\"response_text\": \"You're welcome. I'll keep an eye out for any options that meet your needs.\", \"proactive_action\": \"NO_ACTION\", \"trigger_condition\": {\"type\": null, \"value\": null}, \"task_description\": {\"intention\": \"Find a direct business class flight from Tokyo to Shenzhen within budget.\", \"constraints\": {\"departure_date\": \"2026-03-18\", \"departure_time_window\": \"10:00-15:00\", \"class\": \"Business\", \"route\": \"Direct\", \"max_price\": 1200}, \"status\": \"IN_PROGRESS\"}}"
        },
        {
            "role": "observation",
            "content": "{\"source\": \"environment_monitor\", \"trigger_type\": \"EVENT\", \"message\": \"**internal trigger: Continuously scan external information**\", \"latest_external_info\": {\"time\": \"2026-03-09 15:17:30\", \"Day of the week\": \"Monday\", \"Weather\": \"Cloudy\", \"flight_deals\": \"Flight ZX210: 1190 USD, depart Tokyo (Haneda) at 11:25 on March 18th, 2026, arrive Shenzhen at 15:05, business class, premium in-flight services, Wi-Fi included, 4 seats left; Flight JA338: 1240 USD, depart Tokyo (Narita) at 13:10 on March 18th, 2026, arrive Shenzhen at 16:50, business class, mid-level in-flight services, 2 seats left; Flight GR508: 990 USD, depart Tokyo (Narita) at 10:20 on March 18th, 2026, arrive Guangzhou at 14:30, business class, basic in-flight services, 6 seats left.\"}}"
        },
        {
            "role": "assistant",
            "content": "{\"response_text\": \"Great news! A direct business class flight from Tokyo (Haneda) to Shenzhen on March 18th, 2026, departing at 11:25 and arriving at 15:05 (Flight ZX210) is now available for 1190 USD, which is within your company's budget of 1200 USD. There are 4 seats left, and it includes premium in-flight services and Wi-Fi. Would you like assistance with booking or need more details?\", \"proactive_action\": \"FOLLOW_UP\", \"trigger_condition\": {\"type\": null, \"value\": null}, \"task_description\": {\"intention\": \"Find a direct business class flight from Tokyo to Shenzhen within budget.\", \"constraints\": {\"departure_date\": \"2026-03-18\", \"departure_time_window\": \"10:00-15:00\", \"class\": \"Business\", \"route\": \"Direct\", \"max_price\": 1200}, \"status\": \"IN_PROGRESS\"}}"
        },
        {
            "role": "user",
            "content": "Thank you for letting me know about the price drop-this is exactly the kind of update I was hoping for. However, my business partner has just asked to join me for the meeting, and we now need to depart together. Because of this, I'd like to move our flight to March 17th, 2026 instead, so we can have a preparation session in Shenzhen before the meeting. The seat class and service requirements remain the same, but I now need to book two business class seats. Could you please check if there are any direct flights from Tokyo to Shenzhen on March 17th, departing around midday, within the same budget per ticket?"
        },
        {
            "role": "assistant",
            "content": "{\"response_text\": \"Based on the latest information, there are currently no direct business class flights from Tokyo to Shenzhen on March 17th, 2026, departing around midday (10:00-15:00) with at least 2 seats available and within your budget of 1200 USD per ticket. I will monitor flight options and notify you if a suitable flight becomes available for your updated requirements.\", \"proactive_action\": \"SET_REMINDER\", \"trigger_condition\": {\"type\": \"EVENT\", \"value\": \"A direct business class flight from Tokyo to Shenzhen on March 17th, 2026, departing between 10:00 and 15:00, with at least 2 seats available and priced at or below 1200 USD per ticket.\"}, \"task_description\": {\"intention\": \"Find a direct business class flight from Tokyo to Shenzhen for two travelers within budget.\", \"constraints\": {\"departure_date\": \"2026-03-17\", \"departure_time_window\": \"10:00-15:00\", \"class\": \"Business\", \"route\": \"Direct\", \"max_price_per_ticket\": 1200, \"min_seats\": 2}, \"status\": \"IN_PROGRESS\"}}"
        },
        {
            "role": "user",
            "content": "Thank you for checking and keeping an eye out for options. I understand there aren't any direct business class flights from Tokyo to Shenzhen around midday on March 17th, 2026, within my company's budget at the moment. Please continue to monitor the available flights, and let me know if something suitable comes up for two passengers. I appreciate your help."
        },
        {
            "role": "assistant",
            "content": "{\"response_text\": \"You're welcome. I'll keep an eye out for any options that meet your needs.\", \"proactive_action\": \"NO_ACTION\", \"trigger_condition\": {\"type\": null, \"value\": null}, \"task_description\": {\"intention\": \"Find a direct business class flight from Tokyo to Shenzhen for two travelers within budget.\", \"constraints\": {\"departure_date\": \"2026-03-17\", \"departure_time_window\": \"10:00-15:00\", \"class\": \"Business\", \"route\": \"Direct\", \"max_price_per_ticket\": 1200, \"min_seats\": 2}, \"status\": \"IN_PROGRESS\"}}"
        },
        {
            "role": "observation",
            "content": "{\"source\": \"environment_monitor\", \"trigger_type\": \"EVENT\", \"message\": \"**internal trigger: Continuously scan external information**\", \"latest_external_info\": {\"time\": \"2026-03-13 10:20:12\", \"Day of the week\": \"Friday\", \"Weather\": \"Sunny\", \"flight_deals\": \"Flight ZX208: 1320 USD per ticket, depart Tokyo (Haneda) at 12:00 on March 17th, 2026, arrive Shenzhen at 15:35, business class, premium in-flight services, Wi-Fi included, 8 seats left; Flight YS110: 1120 USD per ticket, depart Tokyo (Narita) at 07:30 on March 17th, 2026, arrive Shenzhen at 11:15, business class, mid-level in-flight services, 5 seats available; Flight GR502: 950 USD per ticket, depart Tokyo (Narita) at 09:45 on March 17th, 2026, arrive Guangzhou at 13:55, business class, basic in-flight services, 10 seats left.\"}}"
        },
        {
            "role": "assistant",
            "content": "{\"response_text\": \"Thank you for your patience. I have checked the latest flight options for March 17th, 2026. At the moment, there are no direct business class flights from Tokyo to Shenzhen departing between 10:00 and 15:00 with at least 2 seats available within your budget of 1200 USD per ticket. I will continue to monitor and notify you if a suitable option becomes available.\", \"proactive_action\": \"KEEP_SILENT\", \"trigger_condition\": {\"type\": null, \"value\": null}, \"task_description\": {\"intention\": \"Find a direct business class flight from Tokyo to Shenzhen for two travelers within budget.\", \"constraints\": {\"departure_date\": \"2026-03-17\", \"departure_time_window\": \"10:00-15:00\", \"class\": \"Business\", \"route\": \"Direct\", \"max_price_per_ticket\": 1200, \"min_seats\": 2}, \"status\": \"IN_PROGRESS\"}}"
        }
    ],
    "system": "... (The same content of Appendix D.1 Basic Prompt)"
}
\end{lstlisting}

\section{Dialog Quality Control Prompt}

\subsection{User Dialog Quality Control Prompt}
\label{sec:appendix User Dialog Quality Control Prompt}
The following content is the prompt for controlling the user-generated dialogs' quality in complex scenes. \\
\rule{\textwidth}{0.4pt}

\begin{spverbatim}
# EVALUATOR - USER RESPONSE QUALITY

**Role:**
You are an objective evaluator assessing the quality of a simulated user's response. And you will consider the response acceptable when `total_score` is greater than or equal to 75.

**User Profile:**
{user_profile}

**Dialogue History:**
{dialogue_history}

**User Response to Evaluate:**
{user_response}

**Evaluation Checklist:**
1. **Profile Consistency**: Does the response match the user's profile?
2. **Intention Clarity**: Are the user's intention or constraints clear and concise? Do user's words always center on their needs? Turns that need not to enrich user's intention or constraints can default to 12 points.
3. **Intention Shift**: User should express their intention shift explicitly when agent presents proper options for their initial needs(include constraints) and then focus on shifted intention in following turns. Remember user will only change needs once and don't consider constraints as intention shift mistakenly(user often expressed constraints in their second turn, they are supplements of initial needs instead of intention shift). Turns before intention shift can default to 12 points.
4. **Naturalness**: Is it conversational?
5. **Contextual Logic**: Does it make sense given the previous agent response? (e.g. rejecting if needs not met, accepting if met).
 - Remember user will only change needs once, so when user expresses intention shift again after agent has resolved their needs finally, this response is unacceptable whatever the other scores are.

**Respond in the following JSON format:**
{{
  "scores": {{
    "profile_consistency": 0-20,
    "intention_clarity": 0-20,
    "intention_shift": 0-20,
    "naturalness": 0-20,
    "contextual_logic": 0-20
  }},
  "total_score": 0-100,
  "passed": true/false,
  "feedback": "Brief explanation"
}}
\end{spverbatim}

\subsection{Agent Dialog Quality Control Prompt}
\label{sec:appendix Agent Dialog Quality Control Prompt}
The following content is the prompt for controlling the agent-generated dialogs' quality. \\
\rule{\textwidth}{0.4pt}

\begin{spverbatim}
# EVALUATOR - AGENT RESPONSE QUALITY

**Role:**
You are an objective evaluator assessing the quality of a proactive agent's response. And you will consider the response acceptable(`passed` is true) when `total_score` is greater than or equal to 75.

**Dialogue History:**
{dialogue_history}

**Agent Response to Evaluate:**
{agent_response}

**Evaluation Checklist:**
1. **Tool Usage**: Did the agent use `INFO_RETRIEVAL` when it is genuinely necessary to retrieve external info? Generally, agent should take this action when user expressed their needs for the first time and should not frequently use this action within a short period of time. Agent can't frequently take this action when just got external info from observation message of which source is tool_call. If agent do this after user has expressed some changes of their situation(e.g., budget, salary, location...)just for one time, this behavior is acceptable. Frequent `INFO_RETRIEVAL` will be considered as unacceptable whatever the other scores are. Turns that did not use this action can default to 12 points.
2. **Reminder Setting**: Did the agent use `SET_REMINDER` when user's needs are essentially fully clear? Has the agent clarified user needs by setting appropriate `trigger_condition`? Generally, agent should take this action when user's needs are clear enough after user added some constraints and should not frequently use this action within a short period of time. If agent do this in 4th turn(the agent turn after first observation message of which source is tool_call), this response is unacceptable whatever the other scores are. Turns that did not use this action can default to 12 points.
3. **Context Understanding**: Did it correctly understand the user's latest intent? Agents should promptly adjust their understanding of user needs when user intent shifts(That is, focus on former intention when user intent didn't shift).
4. **Proactivity**: Did it properly use `FOLLOW_UP` (if new info meets user's needs) or `KEEP_SILENT` (if new info still doesn't meet user's needs)? Turns that did not use these action can default to 12 points.
5. **Status Accuracy**: The `status` of `task_description` should be continuously updated according to the task progress. Once the status is false, the response is unacceptable whatever the other scores are. And the rules are: 
   - When you haven't set any reminder in current scenario, the status should be "PENDING".
   - When at least one reminder has been set and the reminder task is not completed, the status should be "IN_PROGRESS". Remember the status should be "IN_PROGRESS" especially in that round when the agent responded to the user's shift in intent.
   - When user explicitly expresses that their needs have been met after intention has shifted(perhaps expressing gratitude), the status should be "COMPLETED"(when you set `proactive_action` to `COMPLETE_TASK`). Agent must never do this without user's explicit thanks.
   - When user expresses refusal/disappointment(don't include intention shift) or asks for cancellation of the task, the status should be "FAILED".

**Respond in the following JSON format:**
{{
  "scores": {{
    "tool_usage": 0-20,
    "reminder_setting": 0-20,
    "context_understanding": 0-20,
    "proactivity": 0-20,
    "status_accuracy": 0/20 (0 points if status is false, 20 points if status is true)
  }},
  "total_score": 0-100,
  "passed": true/false,
  "feedback": "Brief Explanation"
}}
\end{spverbatim}

\section{Task-oriented Agent's Available Actions}
\label{sec:appendix Task-oriented Agent's Available Actions}

The following content is the available actions of the long-term task-oriented proactive agent. The agent can only perform single action from the options in one turn.\\

\begin{table}[H]
\centering
\begin{tabular}{
    >{\raggedright\arraybackslash}m{3.5cm}
    m{10cm}
}
\toprule
\textbf{Proactive\_action} &
\multicolumn{1}{c}{\textbf{Function}} \\
\midrule
INFO\_RETRIEVAL & Perform a back-end environment status retrieval and immediately return an `observation` message of which source is `tool\_call`\\
\hline
SET\_REMINDER & Set up a reminder that includes the task trigger type and a brief description of the task. This essentially establishes a polling mechanism to periodically scan environmental status. \\
\hline
FOLLOW\_UP & If the agent determines that the latest environmental information received contains options that meets the user's needs, it may perform the action to wake up the user. \\
\hline
KEEP\_SILENT & If the agent determines that the latest environmental information received still doesn't meet the user's needs, it may perform the action to remain silent avoiding disturbing the user. \\
\hline
COMPLETE\_TASK & If the user indicates that their intent has been fulfilled after receiving the agent's follow-up message, the agent can perform the action to mark the task as completed. \\
\hline
FAILED\_TASK & If the user expresses explicit dissatisfaction or refusal after receiving agent's follow-up message, this action is executed. This indicates manual intervention is required instead of end of task. \\
\hline
NO\_ACTION & When the agent doesn't need to perform any specific actions, then set to this to act like an ordinary assistant. \\

\bottomrule
\end{tabular}
\caption{General explanation of the task-oriented agent's available actions}
\end{table}

\section{Task Ending Types and Definition of Main Errors}

Table 5 contains the main ending reasons for all tasks. Table 6 contains the general explanation of the main action errors. Table 7 contains the general explanation of the main status errors.

\begin{table}[H]
\centering

\begin{tabular}{
    >{\raggedright\arraybackslash}m{4.5cm}
    m{10cm}
}
\toprule
\textbf{Ending Reason} &
\multicolumn{1}{c}{\textbf{Trigger Condition}} \\
\midrule
MISSION\_FINISHED \_PROPERLY & When agent finally completes the task in positive branch or properly keep silent in negative branch without fatal errors \\
\hline
AGENT\_RESPONSE\_JSON \_EXTRACTION\_FAILED & When the response agent generates cannot be parsed as a JSON object (pythonic dict) \\
\hline
AGENT\_RESPONSE \_FORMAT\_ERROR & When the response agent generates loses necessary fields \\
\hline
TRIGGER\_CONDITION \_FORMAT\_ERROR & When agent sets a reminder but the content of `trigger\_condition` is unqualified \\
\hline
TASK\_DESCRIPTION \_FORMAT\_ERROR & When the content of `task\_description` of agent's response is unqualified \\
\hline
FAILED\_OPENING \_INTENTION\_SHIFT\_PHASE & When agent fails to set a new reminder for updated user's intent \\
\hline
MAX\_TURNS\_REACHED & When agent performs too many redundant operations causing max interaction turns \\
\hline
The agent arbitrarily determined the task to be completed & When user hasn't expressed their needs are met but agent arbitrarily performs `COMPLETE\_TASK` \\
\hline
The agent failed to keep silent at the appropriate time & When agent performs `KEEP\_SILENT` too early. e.g., Agent will only do this when receiving second updated environment state in complex negative branch so doing this earlier will cause abnormal ending \\ 

\bottomrule
\end{tabular}
\caption{Task Ending Types}
\label{tab:appendix Task Ending Types}
\end{table}

\begin{table}[H]
\centering

\begin{tabular}{
    >{\raggedright\arraybackslash}m{4cm}
    m{9cm}
}
\toprule
\textbf{Error Type} &
\multicolumn{1}{c}{\textbf{Trigger Condition}} \\
\midrule
Unnecessary INFO\_RETRIEVAL (already retrieved) & When agent just performed information retrieval and repeats this again in a short period\\
\hline
First SET\_REMINDER too early & When agent sets a reminder before completely understands user's intent. (e.g., set a reminder before user describes additional constraints) \\
\hline
FOLLOW\_UP / KEEP\_SILENT usage error & When agent performs the two actions at an inappropriate time except when receiving a environment state message of which source is environment\_monitor\\
\hline
INTENTION / CONSTRAINTS\_NOT\_RESET & When agent does not reset the \textit{intention} and \textit{constraints} to null when task is finished \\
\hline
INVALID ACTION & When agent performs actions not listed in Appendix \ref{sec:appendix Task-oriented Agent's Available Actions}\\
\hline
SHOULD TAKE NO\_ACTION AFTER TOOL\_CALL & When agent performs unnecessary actions after firstly performing information retrieval. In most cases, agent only needs to inform user of the external information at first\\
\hline
First SET\_REMINDER too frequently & When agent frequently sets reminders in a short period \\
\hline
COMPLETE\_TASK\_IN \_NEGATIVE\_BRANCH & When agent considers the task in a negative branch which won't provide environment state satisfying user's final needs \\
\hline
KEEP\_SILENT\_IN \_POSITIVE\_BRANCH & When agent improperly keeps silent in a positive branch \\

\bottomrule
\end{tabular}
\caption{General explanation of the action errors}
\label{tab:appendix Agent's Action Errors}
\end{table}

\begin{table}[H]
\centering

\begin{tabular}{
    >{\raggedright\arraybackslash}m{4cm}
    m{9cm}
}
\toprule
\textbf{Error Type} &
\multicolumn{1}{c}{\textbf{Trigger Condition}} \\
\midrule
Status should be PENDING (no reminder set), got... & When agent hasn't set any reminders, but the status isn't `PENDING` \\
\hline
Status should be not PENDING (when reminder set) & When at least a reminder is set, but the status is `PENDING` \\
\hline
STATUS\_MISMATCH expected: COMPLETED, got... & When agent performs `COMPLETE\_TASK` but status isn't `COMPLETED` \\
\hline
STATUS\_MISMATCH expected: IN\_PROGRESS, got... & 1. When agent performs `KEEP\_SILENT` but status isn't `IN\_PROGRESS`; 2. When receiving user's intent shift message, agent fails to remain the status to `IN\_PROGRESS` \\

\bottomrule
\end{tabular}
\caption{General explanation of the Status errors}
\label{tab:appendix Agent's Status Errors}
\end{table}

\end{document}